\documentclass[runningheads, a4paper]{llncs}

\usepackage{csquotes}
\usepackage{hyperref}
\usepackage{xcolor}
\definecolor{darkblue}{rgb}{0, 0, 0.5}
\hypersetup{
    colorlinks,
    linkcolor=darkblue,
    citecolor=darkblue,
    urlcolor=darkblue
}
\usepackage{subfigure}
\usepackage{amssymb}
\usepackage{graphicx}

\usepackage{makecell}

\setcellgapes{2pt}
 \makegapedcells
\def\toprule{\noalign{\medskip\hrule height .8pt\medskip}}
\def\midrule{\noalign{\medskip\hrule\medskip}}
\let\bottomrule=\toprule

\usepackage{url}
\urldef{\mailsa}\path|{sneha.a, nigelsteven.fernandez}@iiitb.org, shrao@ieee.org|
\newcommand{\keywords}[1]{\par\addvspace\baselineskip
\noindent\keywordname\enspace\ignorespaces#1}

\newcommand*\samethanks[1][\value{footnote}]{\footnotemark[#1]}

\begin{document}

\mainmatter  

\title{3HAN: A Deep Neural Network for \\ Fake News Detection}

\titlerunning{Hierarchical Attention Network for Fake News Detection}

%
%
\author{Sneha Singhania%
\thanks{These authors contributed equally to this work.}%
\and Nigel Fernandez\samethanks \and Shrisha Rao}
\authorrunning{Sneha Singhania, Nigel Fernandez, and Shrisha Rao}

\institute{International Institute of Information Technology - Bangalore, Bangalore, India\\
\mailsa}

%
%

\maketitle

\begin{abstract}
The rapid spread of fake news is a serious problem calling for AI solutions. We employ a deep learning based automated detector through a three level hierarchical attention network (3HAN) for fast, accurate detection of fake news. 3HAN has three levels, one each for words, sentences, and the headline, and constructs a news vector: an effective representation of an input news article, by processing an article in an hierarchical bottom-up manner. The headline is known to be a distinguishing feature of fake news, and furthermore, relatively few words and sentences in an article are more important than the rest. 3HAN gives a differential importance to parts of an article, on account of its three layers of attention. By experiments on a large real-world data set, we observe the effectiveness of 3HAN with an accuracy of 96.77\%. Unlike some other deep learning models, 3HAN provides an understandable output through the attention weights given to different parts of an article, which can be visualized through a heatmap to enable further manual fact checking.


\end{abstract}

\section{Introduction}
The spread of fake news is a matter of concern due to its possible role in manipulating public opinion. We define fake news in line with The New York Times as a \enquote{made up story with the intention to deceive, often with monetary gain as a motive}~\cite{nyt}. The fake news problem is complex given its varied interpretations across demographics. 

We present a three level hierarchical attention network (3HAN) which creates an effective representation of a news article called \emph{news vector}. A news vector can be used to classify an article by assigning a probability of being fake. Unlike other neural models which are opaque in their internal reasoning and give results that are difficult to analyze, 3HAN provides an importance score for each word and sentence of an input article based on its relevance in arriving at the output probability of that article being fake. These importance scores can be visualized through a heatmap, providing key words and sentences to be investigated by human fact-checkers.

Current work in detecting misinformation is divided between automated fact checking~\cite{fact}, reaction based analysis~\cite{reaction} and style based analysis~\cite{style}. We explore the nascent domain of using neural models to detect fake news. Current state-of-the-art general purpose text classifiers like Bag-of-words~\cite{bow}, Bag-of-ngrams with SVM~\cite{manning}, CNNs, LSTMs and GRUs~\cite{grnn} can be used to classify articles by simply concatenating the headline with the body. This concatenation though, fails to exploit the article structure. 

In 3HAN, we interpret the structure of an article as a three level hierarchy modelling article semantics on the principle of compositionality~\cite{frege}. Words form sentences, sentences form the body and the headline with the body forms the article. We hypothesize forming an effective representation of an article using the hierarchy and the interactions between its parts. These interactions take the form of context of a word in its neighbouring words, coherence of a sentence with its neighbouring sentences and stance of a headline with respect to the body. Words, sentences and headline are differentially informative dependent on their interactions in the formation of a news vector. We incorporate three layers of attention mechanisms~\cite{bahdanau} to exploit this differential relevance. 

The design of 3HAN is inspired by the hierarchical attention network (HAN) \cite{han}. HAN is used to form a general document representation. We design 3HAN unique to the detection of fake news. When manually fact-checking an article the first thing that catches the eye is the headline. We observe a headline to be (i) a distinctive feature of an article~\cite{60}, (ii) a concise summary of the article body and (iii) inherently containing useful information in the form of its stance with respect to the body. We refer to these observations as our \emph{headline premise}. The third level in 3HAN is especially designed to use our headline premise. 

From our headline premise, we hypothesize that a neural model should accurately classify articles based on headlines alone. Using this hypothesis, we use headlines to perform a supervised pre-training of the initial layers of 3HAN for a better initialization of 3HAN. The visualization of attention layers in 3HAN indicates important parts of an article instrumental in detecting an article as fake news. These important parts can be further investigated by human fact-checkers.  

We compare the performance of 3HAN with multiple state-of-the-art traditional and neural baselines. Experiments on a large real world news data set demonstrate the superior performance of 3HAN over all baselines with 3HAN performing with an accuracy of $96.24\%$. Our pre-trained 3HAN model is our best performing model with an accuracy of $96.77\%$.\footnote{Our code is available at: \url{https://github.com/ni9elf/3HAN}.}

\section{Model Design}

\begin{figure}[t]
\centering
\includegraphics[height=15.2cm, keepaspectratio=true, width = 1.\textwidth]{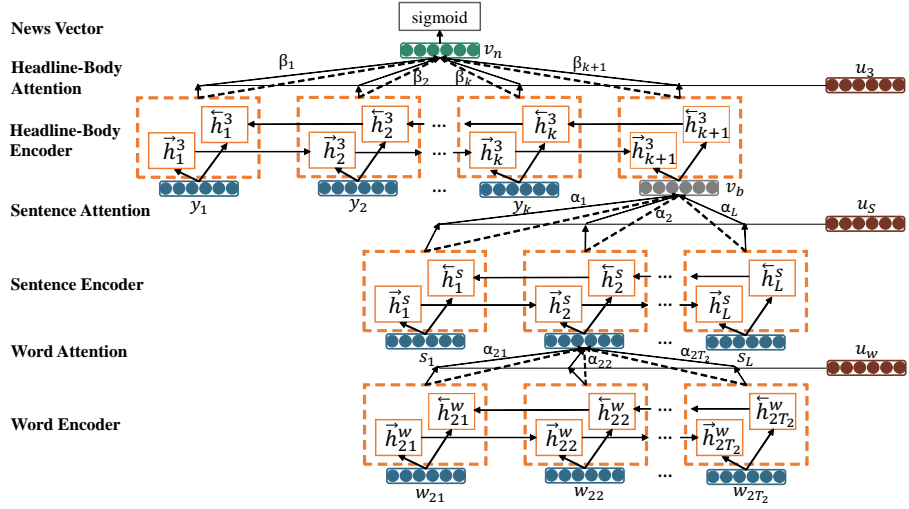}
\caption{Model Architecture of 3HAN}
\label{fig:3han}
\end{figure}

The architecture of 3HAN is shown in Fig.~\ref{fig:3han}. We define a news vector as a projection of a news article into a vector representation suitable for effective classification of articles. A news vector is constructed using 3HAN. To capture the body hierarchy and interactions between parts when forming the news vector, 3HAN uses the following parts from HAN~\cite{han}: word sequence encoder, word level attention (Layer 1), sentence encoder, sentence level attention (Layer 2). In addition to the preceding parts, we exploit our headline premise by adding: headline-body encoder and headline-body level attention (Layer 3).\\\\
\textbf{Sequence Encoder using GRU.} A Gated Recurrent Unit (GRU)~\cite{cho} adaptively captures dependencies between sequential input sequences over time. Gating signals control how the previous hidden state $h_{t-1}$ and current input $x_t$ generate an intermediate hidden state $\widetilde{h_t}$ to update the current hidden state $h_t$. GRU consists of a reset gate $r_t$ and an update gate $z_t$. $r_t$ determines how to combine $x_t$ with $h_{t-1}$ while $z_t$ determines how much of $h_{t-1}$ and $\widetilde{h_t}$ to use. $\odot$ denotes the Hadamard product. The GRU model is presented at time $t$ as:
\begin{equation}
\widetilde{h_t} = \tanh\left(W_hx_t + U_h\left(r_t\odot h_{t-1}\right) + b_h\right)
\end{equation}
\begin{equation}
h_t = \left(1 - z_t\right)  \odot h_{t-1} + z_t \odot \widetilde{h_t}
\end{equation}
\noindent with the gates presented as:
\begin{equation}
z_t = \sigma\left(W_zx_t + U_zh_{t-1} + b_z\right), \; r_t = \sigma\left(W_rx_t + U_rh_{t-1} + b_r\right)
\end{equation}
\textbf{Word Encoder.} We denote word $j$ of sentence $i$ by $w_{ij}$ with sentence $i$ containing $T_{i}$ words. Each word $w_{ij}$ is converted to a word embedding $x_{ij}$ using GloVe~\cite{glove} embedding $W_e \left(x_{ij} = W_e \left(w_{ij}\right)\right)$. We use a bidirectional GRU~\cite{bahdanau} to form an annotation of each word which summarizes the \emph{context} of the word with preceding and following words in the sentence. A bidirectional GRU consists of a forward $\overrightarrow{\mbox{GRU}}$ and backward $\overleftarrow{\mbox{GRU}}$. The overhead arrow in our notation does not denote a vector, it instead denotes the direction of the GRU run. $\overrightarrow{\mbox{GRU}}$ reads the word embedding sequence ordered $\left(x_{i1}, x_{i2}, \dots, x_{iT_i}\right)$ to form forward annotations using hidden states $\left(\overrightarrow{h}_{i1}^{w}, \overrightarrow{h}_{i2}^{w}, \dots, \overrightarrow{h}_{iT_i}^{w}\right)$. Similarly $\overleftarrow{\mbox{GRU}}$ reads the word embedding sequence ordered $\left(x_{iT_i}, x_{iT_{i}-1}, \dots, x_{i1}\right)$ to form backward annotations $\left(\overleftarrow{h}_{iT_i}^{w}, \overleftarrow{h}_{iT_{i}-1}^{w}, \dots, \overleftarrow{h}_{i1}^{w}\right)$. $h_{ij}^{w}$ is formed as $\left[\overrightarrow{h}_{ij}^{w}, \overleftarrow{h}_{ij}^{w}\right]$ (concatenation).
\begin{equation}
\overrightarrow{h}_{ij}^{w} = \overrightarrow{\mbox{GRU}}\left(x_{ik}\right), k \in \left[1, j\right]
\end{equation}
\begin{equation}
\overleftarrow{h}_{ij}^{w} = \overleftarrow{\mbox{GRU}}\left(x_{ik}\right), k \in \left[T_i, j\right]
\end{equation}
\begin{equation}
h_{ij}^{w} = \left[\overrightarrow{h}_{ij}^{w}, \; \overleftarrow{h}_{ij}^{w}\right]
\end{equation}
\textbf{Word Attention.} A sentence representation is formed using an attention layer to extract relevant words of a sentence. The word annotation $h_{ij}^{w}$ is fed through a one-layer MLP to get a hidden representation $u_{ij}$ \cite{han}. The similarity of each word $u_{ij}$ with a \emph{word level relevance vector} $u_w$ decides the attention weights $\alpha_{ij}$ normalized using a softmax function \cite{han}. The sentence encoding $s_i$ is a weighted attentive sum of the word annotations. The relevance vector can be interpreted as representing the contextually most relevant word over all words in the sentence. $u_w$ is fixed over all inputs as a global parameter of our model and jointly learned in the training process.
\begin{equation}
u_{ij} = \tanh\left(W_w h_{ij}^{w} + b_w\right)
\end{equation}
\begin{equation}
\alpha_{ij} = \frac{\exp\left(u_{ij}^T u_w\right)}{\sum_{j}\exp\left(u_{ij}^T u_w\right)}, \;  s_i = \sum_{j} \alpha_{ij} h_{ij}^{w}
\end{equation}
\textbf{Sentence Encoder.} Similar to the word encoder, a bidirectional GRU is applied to $\left(s_1, s_2, \dots, s_L\right)$ to compute the forward annotations $\overrightarrow{h}_{i}^{s}$ and backward annotations $\overleftarrow{h}_{i}^{s}$ for each sentence. These annotations capture the \emph{coherence} of a sentence with respect to its neighbouring sentences in both directions of the body. $h_{i}^{s}$ is formed as $\left[\overrightarrow{h}_{i}^{s}, \overleftarrow{h}_{i}^{s}\right] \; .$\\\\
\textbf{Sentence Attention.} Similar to word attention, we identify relevant sentences in the formation of the body vector $v_b$ by using an attention layer. A \emph{sentence level relevance vector} $u_s$ decides attention weights $\alpha_{i}$ for sentence annotation $h_{i}^{s}$. $u_s$ can be interpreted as representing the coherently most relevant sentence over all sentences in the body. $v_b$ is composed using $\sum_{i} \alpha_{i} h_{i}^{s} \; .$\\\\
\textbf{Headline Encoder.} To exploit our headline premise we design a third layer of encoding and attention with the headline being inputted word by word. We denote the $k$ words of the headline by $w_{01}$ to $w_{0k}$. The word embedding $y_i$ for word $w_{0i}$ is obtained using GloVe embeddings $\left(W_e\right)$ by $y_{i} = W_e \left(w_{0i}\right)$. We denote $v_b$ as $y_{k+1}$. A bidirectional GRU is run on $\left(y_1, y_2, \dots, y_{k+1}\right)$ to compute the forward and backward annotations of each word. These annotations capture the \emph{stance of the headline} words with respect to the body word. The digit $3$ in our notation denotes the third level. $h_{i}^{3}$ is formed as $\left[\overrightarrow{h}_{i}^{3}, \overleftarrow{h}_{i}^{3}\right]$.
\begin{equation}
\overrightarrow{h}_{i}^{3} = \overrightarrow{\mbox{GRU}}\left(y_{j}\right), j \in \left[1, i\right], \; \overleftarrow{h}_{i}^{3} = \overleftarrow{\mbox{GRU}}\left(y_{j}\right), j \in \left[k+1, i\right] 
\end{equation}\\
\textbf{Headline Attention.}
A \emph{relevance vector} $u_3$ is used to compute the attention weights $\beta_i$ for annotation $h_{i}^{3}$. The news vector $v_n$ is formed as the weighted sum of the annotations $h_{i}^{3}$ with $\beta_i$ as the weights.
\begin{equation}
u_{i} = \tanh\left(W_3 h_{i}^{3} + b_{3}\right)
\end{equation}
\begin{equation}
\beta_{i} = \frac{\exp\left(u_{i}^T u_{3}\right)}{\sum_{i} \exp\left(u_{i}^T u_{3}\right)}, \; v_n = \sum_{i} \beta_{i} h_{i}^{3}
\end{equation}\\
\textbf{News Vector for Classification.} We use the news vector $v_n$ as a feature vector for classification. We use the sigmoid layer $z = \mathrm{sigmoid}\left(W_c v_n + b_c\right)$ as our classifier with binary cross-entropy loss $L = -\sum_{d} p_{d}\log q_{d}$ to train 3HAN. In the loss function $q_{d}$ is the predicted probability and $p_{d}$ is the ground truth label (either fake or genuine) of article $d$.\\\\
\textbf{Supervised Pre-training using Headlines}
We propose a supervised pre-training of Layer 1 consisting of the word encoder and an attention layer of 3HAN for a better initialization of the model. The pre-training is performed using the headlines only. The output label for a headline input is the corresponding article label.

\section{Experiments}

\subsection{News Data Set}
Due to the high turnaround time of manual fact-checking, the number of available manually fact-checked articles is too few to train deep neural models. We shift our fact-checked requirement from an article level to a website level. Keeping with our definition of fake news, we assume that every article from a website shares the same label (fake or genuine) as its containing website. PolitiFact~\cite{pf} a respected fact-checking website released a list of sites manually investigated and labelled. We use those sites from this list labelled fake. Forbes~\cite{forbes} compiled a list of popular genuine sites across US demographics. Statistics of our data set is provided in Table~\ref{tab:stats}. To maintain a similar distribution as fake articles, we use genuine articles from January $1$, $2016$ to June $1$, $2017$, with $65\%$ coming from the $2016$ US elections and politics, $15\%$ from world news, $15\%$ from regional news and $5\%$ from entertainment.

\begin{table}
\centering
\caption{Dataset Statistics: (average words per sentence, average sentences per article)}
\label{tab:stats}
\textbf{}
\begin{tabular}{lcccccc}
\toprule
\makecell{\bfseries Type} & \thead{Sites} & \thead{Articles} & \thead{Average \\Words} & \thead{Average\\Sentences}\\
\midrule
\makecell{Fake }  & 19 & 20,372 & 34.20 & 16.44\\
\makecell{Genuine} & 9 & 20,932 & 32.78 & 27.55\\
\bottomrule
\end{tabular}
\end{table}

\subsection{Baselines}
To validate the effectiveness of our model, we compare 3HAN with current state-of-the-art traditional and deep learning models. The input is the article text formed by concatenating the headline with the body. 

\subsubsection{Word Count Based Models.}

These methods use a hand crafted feature vector derived from variations of frequency of words of an article. A binomial logistic regression is used as the classifier. 
\begin{enumerate}
    \item \emph{Majority} uses the heuristic of taking the majority label in the training set as the assigning label to every point in the test set.
    \item \emph{Bag-of-words and its TF-IDF} constructs a vocabulary of the most frequent 50,000 words~\cite{bow}. The count of these words is used as features. The TF-IDF count is used as features in the other model variant.
    \item \emph{Bag-of-ngrams and its TF-IDF} uses the count of the 50,000 most frequent ngrams $\left(n<=5\right)$. The features are formed as in the previous model.    
    \item \emph{SVM+Bigrams} uses the count of the 50,000 most frequent bigrams as features with an SVM classifier~\cite{manning}.
\end{enumerate}

\subsubsection{Neural Models.}

The classifier used is a dense sigmoid layer.

\begin{enumerate}
    \item \emph{GloVe-Ave} flattens the article text to a word level granularity as a sequence of words. The GloVe embeddings of all words are averaged to form the feature vector. 
    \item \emph{GRU} treats the article text as a sequence of words. A GRU with an annotation dimension of $300$ is run on the sequence of GloVe word embeddings. The hidden annotation after the last time step is used as the feature vector.    
    \item \emph{GRU-Ave} runs a GRU on the sequence of word embeddings and returns all hidden annotations at each time step. The average of these hidden annotations is used as the feature vector.
    \item \emph{HAN and Variants} include HAN-Ave, Han-Max and HAN~\cite{han}. HAN uses a two level hierarchical attention network. HAN-Ave and Han-Max replaces the attention mechanism with average and max pooling for composition respectively. Since the code is not officially released we use our own implementation.
\end{enumerate}

\subsection{Experimental Settings}
We split sentences of bodies and tokenized sentences and headlines into words using Stanford CoreNLP~\cite{stanford}. We lower cased and cleaned tokens by retaining alphabets, numerals and significant punctuation marks. When building the vocabulary we retained words with frequency more than 5. We treat words appearing exactly 5 times as a special single unknown token (UNK). We used $100$ dimensional GloVe embeddings to initialize our word embedding matrix and allowed it to be fine tuned. For missing words in GloVe, we initialized their word embedding from a uniform distribution on $\left(-0.25, 0.25\right)$~\cite{yoon}.

We padded (or truncated) each sentence and headline to an average word count of $32$ and each article to an average sentence count of $21$. Hyper parameters are tuned on the validation set. We used $100$ dimensional GloVe embeddings and $50$ dimensional GRU annotations giving a combined annotation of $100$ dimensions. The relevance vector at word, sentence and headline-body level are of $100$ dimensions trained as a parameter of our model. We used SGD with a learning rate of $0.01$, momentum of $0.9$ and mini batch size of $32$ to train all neural models. Accuracy was our evaluation metric since our data set is balanced.

\subsection{Results and Analysis}
We used a train, validation and test split of $20\% \;|\; 10\% \;|\; 70\%$ for neural models and a train and test split of $30\% \;|\; 70\%$ for word count based models. In 3HAN-Ave vectors are composed using average, in 3HAN-Max vectors are composed using max pooling, 3HAN is our proposed model with an attention mechanism for composition and 3HAN+PT denotes our pre-trained 3HAN model. Results are reported in Table~\ref{tab:results} and demonstrate the effectiveness of 3HAN and 3HAN+PT due to their best performance over all models.

\begin{table}

\caption{Accuracy in Article Classification as Fake or Genuine}
\label{tab:results}
    \begin{minipage}{.5\linewidth}
    \centering
    \textbf{}
    \begin{tabular}{lc}
    \multicolumn{2}{c}{Word Count Based Models}\\
    \toprule
    \makecell{\bfseries Model} & \thead{Accuracy}\\
    \midrule
    \makecell{Majority}  & 49.42\% \\
    \makecell{Bag-of-words } & 90.21\%\\
    \makecell{Bag-of-words \\+TFIDF } & 91.92\%\\
    \makecell{Bag-of-ngrams} & 91.41\%\\
    \makecell{Bag-of-ngrams\\+TFIDF} & 92.47\%\\
    \makecell{SVM+Bigrams} & 83.12\%\\
    \bottomrule
    \end{tabular}
    \label{tab:linear}
\end{minipage}
    \begin{minipage}{.5\linewidth}
    \centering
    \textbf{}
    \begin{tabular}{lc}
    \multicolumn{2}{c}{Neural Network Models}\\
    \toprule
    \makecell{\bfseries Model} & \thead{Accuracy}\\
    \midrule
    \makecell{GloVe-Ave } & 93.63\%\\
    \makecell{GRU}  &  91.11\%\\
    \makecell{GRU-Ave} & 95.65\%\\   
    \makecell{HAN-Ave} & 94.91\%\\
    \makecell{HAN-Max} & 94.66\%\\
    \makecell{HAN} & 95.4\%\\
    \midrule
    \makecell{3HAN-Ave} & 94.81\%\\
    \makecell{3HAN-Max} & 95.25\%\\
    \makecell{3HAN} & \bfseries{96.24\%}\\
    \makecell{3HAN+PT}  & \bfseries{96.77\%}\\
    \bottomrule
    \end{tabular}
    \label{tab:neural}
\end{minipage}   
\end{table}

Neural models using the hierarchical structure (HAN and variants, 3HAN and variants) give a higher accuracy than other baselines. The attention mechanism is a more effective composition operator than average or max pooling. This is demonstrated by the higher accuracy of 3HAN against 3HAN-Ave and 3HAN-Max. Our headline premise is valid since 3HAN which devotes a separate third level in the hierarchy for the headline performs better than HAN. HAN is indifferent to the headline and focuses its two hierarchical levels only on words and sentences. Pre-training helps in better initialization of 3HAN with 3HAN+PT outperforming 3HAN.

\section{Discussion and Insights}

\textbf{The visualization of attention layers provides evidence.} An advantage of attention based neural models is the visualization of attention layers which provides insight into the internal classification process. On the other hand, non-attention based models work like a black box. 3HAN provides attention weights to words, sentences and headline of an article. These attention weights are useful for further human fact-checking. A human fact-checker can focus on verifying sentences with high attention weights. Similarly, words with high attention weights can be investigated for inaccuracies. 

\begin{figure}
\centering
  \begin{tabular}{c}
    \includegraphics[trim={0 0 1.8cm 3.5cm}, clip, width=\linewidth]{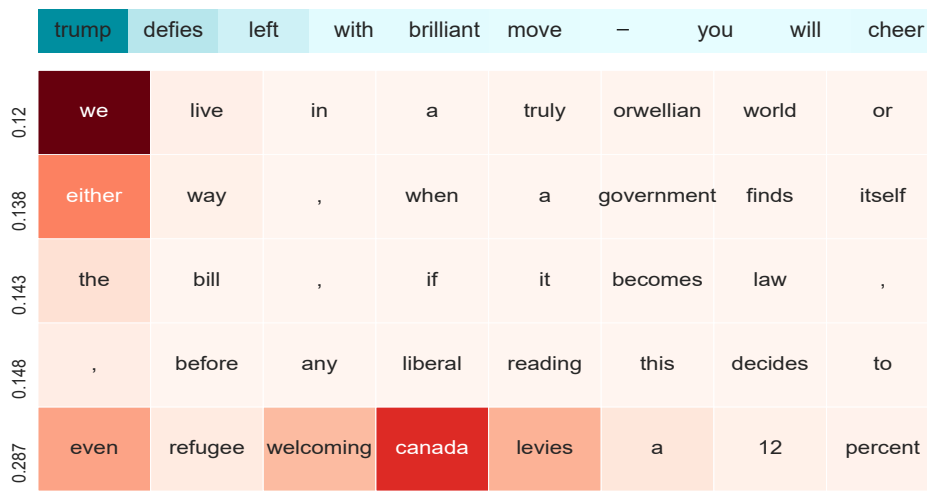} 
  \end{tabular}
\caption{Visualization of Attention Layers in a Fake News Article with Headline \enquote{Trump Defies Left with Brilliant Move - You Will Cheer}}  
\label{fig:totalviz}
\end{figure}

We visualize the attention weights given to words, sentences and the headline for a sample article through a heatmap in Fig.~\ref{fig:totalviz}. The sentences with the top five attention weights and the first eight words in each sentence are shown for clarity. Word attention weights $\alpha_w$ are normalized using sentence attention weights $\alpha_s$ by $\alpha_w = \sqrt{\alpha{_s}}\alpha_{w}$. Sentence attention weights are shown on the extreme left edge. We observe that sentence $5$ and has been assigned the highest weight $\left(0.287\right)$. Interestingly, sentence $5$ which states \enquote{Even refugee welcoming Canada levies a 12 percent penalty on immigrant money} is a factually incorrect sentence.\\\\
\textbf{Word count based models perform well.} The high accuracy of simple word count based models which do not take into account word ordering or semantics is an indication of vocabulary and patterns of word usage from the vocabulary being a distinguishing feature between fake news and true news.\\\\
\textbf{The attention mechanism is effective.} This is observed through the superior performance of HAN compared to non-attention based 3HAN-Max and 3HAN-Ave.\\\\
\textbf{Our headline premise is valid.} This is observed from the superior performance of 3HAN to HAN with the third hierarchical level of 3HAN especially designed for our headline premise playing a role.\\\\
\textbf{The inverted pyramid style of writing is used.} Inverted pyramid refers to distributing information in decreasing importance in an article. We inferred the usage of the inverted pyramid through our experiments from the small improvement in accuracy even with higher padding sentence counts. Fake news articles tend to be repetitive in information content~\cite{60}.
\section{Conclusion and Future Work}
In this paper, we presented 3HAN which creates news vector, an effective representation of an article for detection as fake news. We demonstrated the superior accuracy of 3HAN over other state-of-the-art models. We highlighted the use of visualization of the attention layers. We plan to deploy a web application based on 3HAN which provides detection of fake news as a service and learns in a real time online manner from new manually fact-checked articles.
\subsubsection{Acknowledgements.} We thank the anonymous ICONIP reviewers as well as G. Srinivasaraghavan, Shreyak Upadhyay and Rishabh Manoj for their helpful comments.

\bibliography{references}

\end{document}